\documentclass[12pt]{spieman}  
\usepackage{amsmath,amsfonts,amssymb}
\usepackage{graphicx}
\usepackage{setspace}
\usepackage{tocloft}
\usepackage{url}

\usepackage[utf8]{inputenc}

\title{Three-dimensional multimodal medical imaging system based on free-hand ultrasound and structured light}

\author[a]{Jhacson Meza}
\author[a]{Sonia H. Contreras-Ortiz}
\author[b]{Lenny A. Romero}
\author[a,*]{Andres G. Marrugo}
\affil[a]{Facultad de Ingeniería, Universidad Tecnológica de Bolívar, Cartagena, Colombia}
\affil[b]{Facultad de Ciencias Básicas, Universidad Tecnológica de Bolívar, Cartagena, Colombia}

\cftpagenumbersoff{figure}
\cftpagenumbersoff{table}
\begin{document}
\sloppy
\maketitle

\begin{abstract}
We propose a three-dimensional (3D) multimodal medical imaging system that combines freehand ultrasound and structured light 3D reconstruction in a single coordinate system without requiring registration. To the best of our knowledge, these techniques have not been combined before as a multimodal imaging technique. The system complements the internal 3D information acquired with ultrasound, with the external surface measured with the structure light technique. Moreover, the ultrasound probe's optical tracking for pose estimation was implemented based on a convolutional neural network. Experimental results show the system's high accuracy and reproducibility, as well as its potential for preoperative and intraoperative applications. The experimental multimodal error, or the distance from two surfaces obtained with different modalities, was 0.12~mm. The code is available as a Github repository. 
\end{abstract}


{\noindent \footnotesize\textbf{*}Corresponding author:  \linkable{agmarrugo@utb.edu.co} }

\begin{spacing}{1}
\section{Introduction}
Conventional medical procedures rely heavily on the physician's visualization skills and expertise. These can be enhanced by modern multimodal imaging technologies to perform complex tasks, such as surgical planning, navigation, and intraoperative decision making~\cite{Mascagni:2018fi}. However, most intraoperative systems still rely on two-dimensional (2D) information, although the task is essentially three-dimensional (3D). Moreover, the existing multimodal 3D imaging technologies are either too expensive and bulky to be used in routine procedures~\cite{vanBeek:2019bh} or not free of ionizing radiation, as in the case of Computed Tomography~(CT)~\cite{Mascagni:2018fi}.

There are suitable alternatives, such as Ultrasound (US) imaging, which is free of ionizing radiation, portable, low-cost, and allows real-time visualization~\cite{Contreras:2012}. Furthermore, freehand scanning with a US probe provides a flexible and convenient tool for clinicians to reconstruct 3D images of the region of interest~\cite{Huang:2017cd, colley2017methodology}. However, 3D US does not provide information about the external anatomy, such as that provided by CT or magnetic resonance imaging (MRI), which is essential for surgical planning and navigation.

Recently, digital light processing (DLP) projectors have enabled fast and accurate structured light (SL) systems for 3D surface reconstruction~\cite{Zhang:2018jb}. Although primarily used in the manufacturing industry~\cite{Marrugo:2020fq}, DLP-based SL systems have found new uses in biomedical applications such as in surgical navigation~\cite{Zhang:2019kx}, real-time 3D measurement otoscopy~\cite{VanderJeught:2017jq}, and fluorescence imaging and topography for intraoperative imaging~\cite{quang2017fluorescence}. It is by far the most adequate technique for obtaining a digital representation of the external surface. Moreover, to the best of our knowledge, freehand US and SL have not been used simultaneously as a multimodal imaging technique. It is the programmable nature of the DLP-based SL system that opens new possibilities for such a multimodal technique with freehand US. This combination can be useful for percutaneous interventions such as biopsies, ablations, or drainages that are mainly guided using 2D imaging techniques, specifically with US, due to its versatility~\cite{anas2018deep,anzidei2017imaging}. For these percutaneous procedures, it would be helpful to have the 3D model of the external anatomy to improve image interpretation and spatial understanding of the US images.

In this paper, we propose a multimodal imaging system based on freehand US and SL to acquire the internal and external features of a zone of interest in the same coordinate system. The proposed system is aimed mostly for percutaneous interventions as a preoperative and intraoperative imaging tool. In this way, our system can be used for treatment planning, intervention planning with preoperative data, and procedure guidance due to its ability to easily update previously acquired data. Furthermore, it may provide a low-cost solution for surgical planning in rural or low-resource settings~\cite{Bowden:20}. Our approach considers optical tracking of the US probe and the SL 3D reconstruction using a pair of cameras and a DLP projector, all calibrated to a global coordinate system. Therefore, no registration is needed for multimodal visualization, an often challenging procedure~\cite{cherry2009multimodality}. Moreover, we developed a deep learning US probe tracking algorithm for precise tracking under varying illumination conditions and motion blur. The experimental results show the potential of the proposed system as a multimodal medical imaging technique and a surgical navigation tool.

\section{Related Work}
US and SL are powerful, cost-effective imaging techniques that offer great flexibility and potential solutions to many medical procedures that require precise positioning and guidance. US is highly used for both open and minimally invasive procedures with an array of different probes and newly developed contrast agents~\cite{walker2017intraoperative}. Because of many of these advantages, US imaging has been combined with other techniques. For example, with fluorescence molecular tomography~\cite{li2014ultrasound}, with MRI for intraoperative guidance~\cite{nagelhus2006computer,lindseth2003multimodal}, with fluorescence lifetime imaging for oral cancer evaluation~\cite{fatakdawala2013multimodal}, and intravascular photoacoustic imaging for characterization of atherosclerosis~\cite{li2020multimodal}. In summary, US is used so often with other modalities because it provides the clinicians with interactive feedback~\cite{mela2021novel}. 

Similarly, SL has also been combined with other techniques to take advantage of its ability to probe the surface of the region of interest. For example, it has been combined with fluorescence imaging to obtain a 3D fluorescence topography in which the DLP projector is used as an excitation source~\cite{quang2017fluorescence}. There are other instances in which SL has been used in a multimodal system, like with micro-CT for evaluating excised breast tissue~\cite{mcclatchy2017calibration}, and with a PET system for head motion tracking~\cite{olesen2011motion}. Its capacity for accurate, fast 3D surface reconstruction is unparalleled with competing techniques. 

There are two cases where US and SL systems have been used separately to address the same problem: scoliosis evaluation and breast cancer detection. In the first case, Pino et al.~\cite{pino2016quantification} measured the topographic changes in the back surface obtained through an SL system to monitor idiopathic scoliosis. For the same type of assessment, Cheung et al.~\cite{cheung2015freehand} developed a 3D freehand ultrasound system to measure spine curvature based on bony landmarks of the spine extracted from a stack of US scans. In the second case, US is known to be an excellent technique for breast cancer detection~\cite{vairavan2017brief, vsroubek2019computer}. Moreover, there is some evidence obtained with SL systems linking breast surface variation to cancer~\cite{norhaimi2019breast}. These examples give some insight into why it would be beneficial to have a multimodal freehand US and SL system.

There have been several approaches in which US and SL have been jointly used to improve or solve orientation or image guidance problems in medical imaging, although none so far as a multimodal imaging technique. For example, Horvath et al.,~\cite{horvath2011towards} used an SL system composed of a camera and two lasers attached to a US probe to determine the orientation of the surface relative to the US transducer for solving the Doppler ambiguity problem from arteries and veins running parallel to the surface. Basafa et al.,~\cite{basafa2017visual} used a stereo vision system mounted on the US transducer to aid an US and computer tomography multimodal system for image guidance. Similarly, there have been many attempts at mounting a camera on the US probe for sensorless freehand US~\cite{sun2014probe, wang2017ultrasound}. However, the implementation is not problem-free due to insufficient tracking features in the skin or surrounding regions--not to mention the bulkier US probe with additional instruments onboard.

Finally, US and SL techniques have often been used to solve similar problems in medical imaging. However, considering that they provide complementary information (US internal features and SL external surface), we find it necessary to develop a system with both techniques highly integrated as a multimodal imaging system. It is worth noting that the DLP projector in the SL system enables features such as active surgical guidance and telementoring.

\section{Principles}
\subsection{3D Freehand Ultrasound}
\begin{figure}[b]
    \centering
    \includegraphics[width=0.5\textwidth]{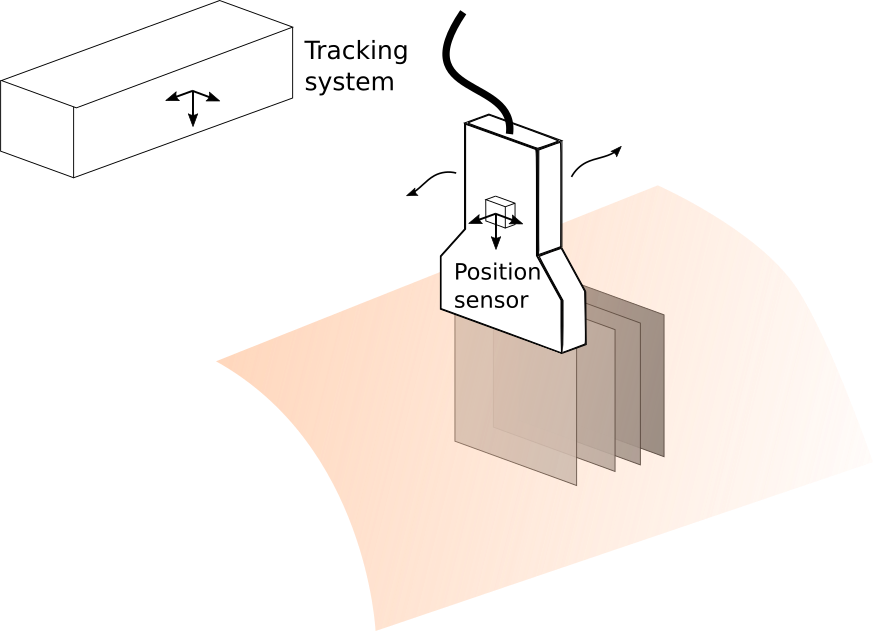}
    \caption{3D freehand ultrasound (US) technique: a tracking system is used to store the probe's pose along with the B-scan. As a result, the 2D US images can be mapped to a 3D volume.}
    \label{fig:us_theory}
\end{figure}

3D freehand US consists of acquiring 2D US images (B-scans) and simultaneously tracking the position and orientation of a probe in space with a position sensor or an optical or electromagnetic tracker system, as shown in Fig.~\ref{fig:us_theory}. Thus, knowing the probe's position and the rigid-body transformation from the US image plane to the probe is sufficient to locate the B-scans in 3D space. The global coordinate system is usually the tracker frame. We can estimate the transformation between the image plane and the probe through a calibration procedure using a phantom, i.e., an object of known geometry. Many phantoms have been proposed in the literature, but the simplest is a point target~\cite{hsu2008comparison} scanned from different positions and orientations. We can build it with cross-wires, a spherical object, or the tip of a stylus. The point phantom of two cross-wires is easy to build and produces accurate calibration results~\cite{prager1998rapid,mercier2005review,Torres:2008kt}.

\subsection{Structured Light}
\begin{figure}
    \centering
    \includegraphics[width=0.5\textwidth]{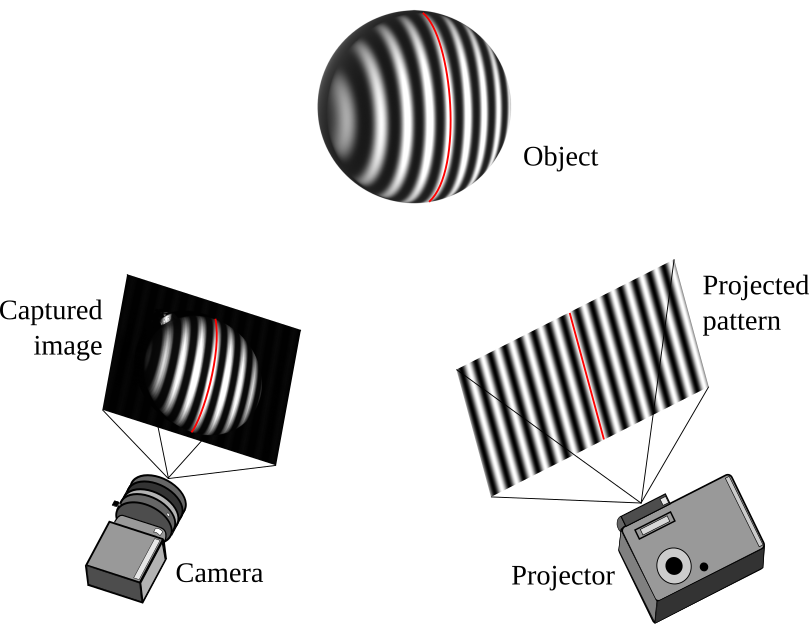}
    \caption{A structured light (SL) setup composed of a camera-projector pair. The projected pattern is distorted due to the object's topography. The red line shows how the pattern is deformed and how the camera detects it.}
    \label{fig:sl_theory}
\end{figure}

Optical 3D shape measurement based on SL is a well-known optical metrology technique with many applications~\cite{Marrugo:2020fq}. Its basic configuration is that of a camera-projector pair, as shown in Fig.~\ref{fig:sl_theory}. The projector projects a structured pattern (typically a sinusoidal pattern) onto the object's surface. It is distorted due to the topographic variations (indicated by a red line). This pattern is captured with a camera and processed to recover the depth information. Phase-shifting is one of the most used phase-retrieval algorithms since it produces pixel-wise phase-maps using at least three patterns~\cite{lu2021motion}. The obtained phase map is wrapped in the range from $-\pi$ to $\pi$. Therefore, an unwrapping procedure is necessary to estimate a continuous phase-map. Finally, with an unwrapped phase distribution, the projector-camera pixel correspondence is solved to perform triangulation and obtain the 3D surface~\cite{juarez2019key}. This approach assumes the SL system has been previously calibrated as a stereo vision system, in which the projector is modeled as an inverse camera~\cite{zhang2006novel}.

\section{Method}
\begin{figure}
    \centering
    \includegraphics[width=0.5\textwidth]{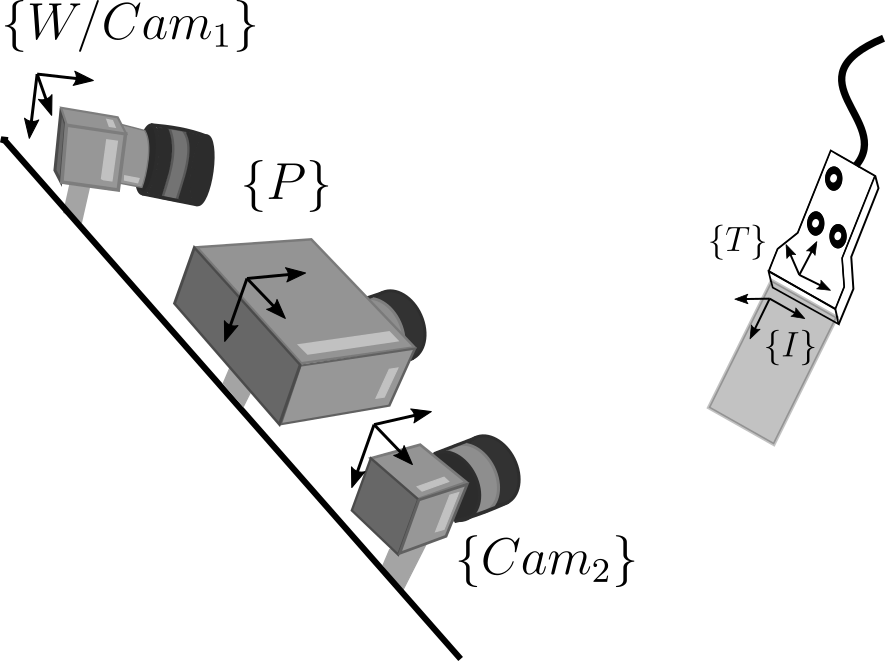}
    \caption{Mulimodal 3D imaging system composed of two cameras, a projector and a US machine.}
    \label{fig:system}
\end{figure}

Our multimodal system shown in Fig.~\ref{fig:system} consists of two monochromatic cameras Basler acA1300-200um (1280x1024, 203 fps), a DLP projector Dell M318WL (1280x800), and a B-mode ultrasound machine Biocare iS 20. The figure shows the main coordinate frames involved in the proposed multimodal imaging method. The world coordinate system $\{W\}$ is placed on the camera 1 frame $\{Cam_1\}$, and the remaining components are referred to this system. The pair camera-1 and projector form the SL system. Whereas camera-1 and 2 form the stereo vision system used for tracking the US probe, which is the basis of the freehand US system. It worth noting that, through the acquisition of both techniques, we do not change the position of the $\{W\}$ frame. Therefore, we avoid registering images for merging the multimodality data, as both 3D data are referenced to the same coordinate system.

\begin{figure}[b]
    \centering
    \includegraphics[width=1\textwidth]{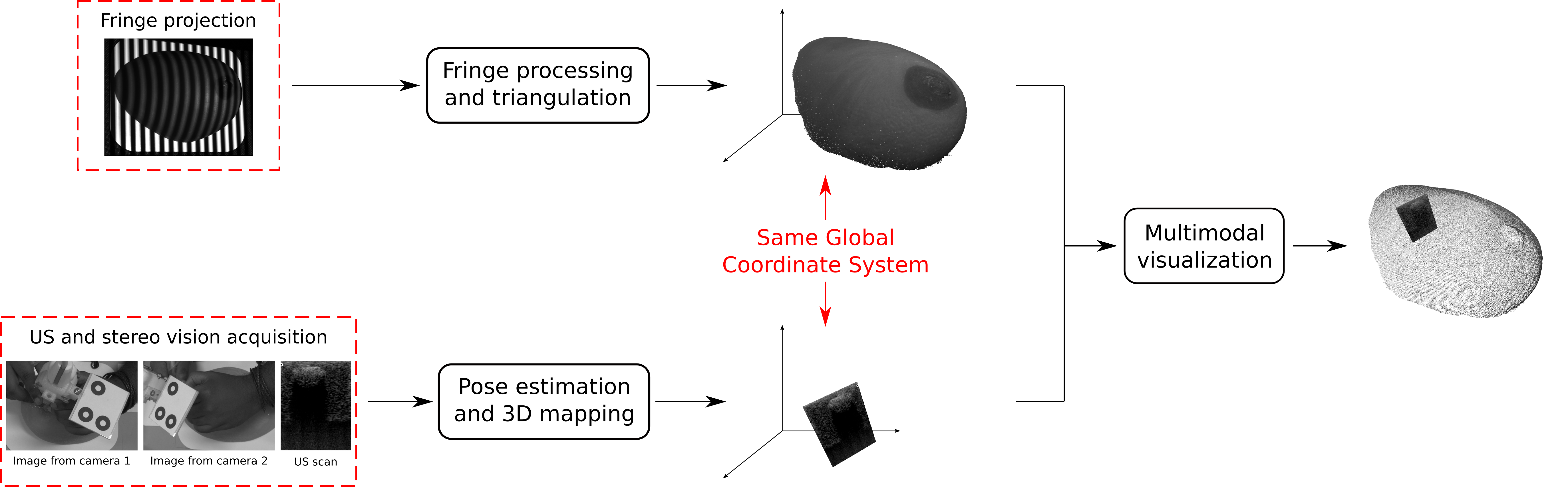}
    \caption{The proposed mulimodal 3D imaging pipeline with the external 3D surface and internal US images.}
    \label{fig:diagram}
\end{figure}

The proposed 3D multimodal imaging pipeline is described in Fig.~\ref{fig:diagram}. On the one hand, to obtain the 3D surface with the SL system, the projector projects vertical patterns onto the object's surface, which camera 1 captures. Using temporal phase unwrapping with the phase-shifting + gray coding technique, we recover the absolute phase map. With the absolute phase map, the camera-projector matches are established, and the object's topography is obtained through triangulation. On the other hand, for the 3D freehand US system, we use a strategy for tracking the probe's pose based on Deep Learning and using the stereo vision system $\{Cam_1\}$ and $\{Cam_2\}$ shown in Fig.~\ref{fig:system}. With an in-house developed acquisition software, we simultaneously acquire the US and camera images. 

For tracking, a target with three coplanar circles was attached to the probe with a 3D-printed piece, as shown in Fig.~\ref{fig:diagram}. Then, with a Convolutional Neural Network (CNN) model, we estimate the three circles' centers with sub-pixel resolution. The network also solves the matching problem between the two views. With the previous data, we estimate the position and orientation of the ultrasound probe. Afterward, the acquired US images are mapped to a 3D space by knowing the transformation between the US image frame $\{I\}$ and the target frame $\{T\}$, which is established through a calibration procedure. Finally, both 3D reconstructions are visualized in the same coordinate frame.

\subsection{Pose estimation procedure}
Recent freehand US systems use dedicated costly optical tracking systems which cannot be easily merged with complementary imaging techniques due to proprietary software and calibration limitations~\cite{colley2017methodology,hu2017freehand}. Alternative optical tracking methods with fiducial markers typically work well under controlled environments. In a previous version of this work~\cite{Meza:2020ge}, we developed a target tracking method based on OpenCV using classical computer vision techniques. Although this method yields accurate tracking and pose estimation results, it often required tuning many parameters to make it work under different environments. For this reason, in this paper, we use MarkerPose~\cite{meza2021markerpose}, a low-cost, real-time pose estimation method based on stereo vision and deep learning.

\begin{figure}
    \centering
    \includegraphics[width=0.4\textwidth]{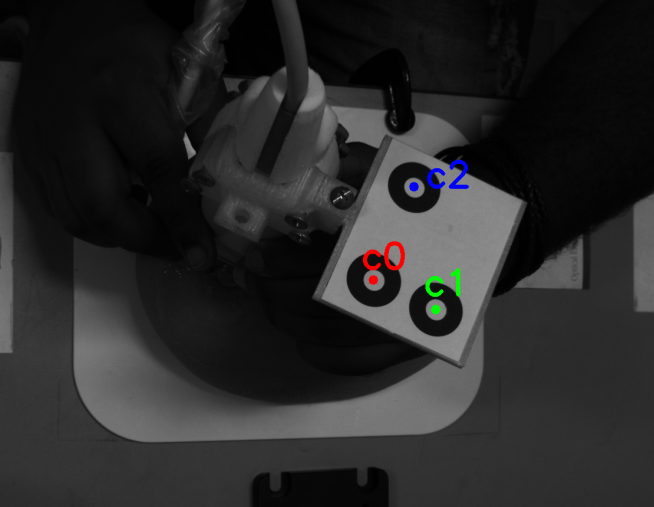}
    \caption{The target with the corresponding circle centers' IDs.}
    \label{fig:target_labels}
\end{figure}

Our three-circles target and their ID labels are used for pose estimation, as shown in Fig.~\ref{fig:target_labels}. With the centers' 3D position of the three coplanar circles, we define a coordinate system, where $c_0$ is the origin, and the vectors $\overrightarrow{c_0c_1}$ and $\overrightarrow{c_0c_2}$ represent the $x$-axis and $y$-axis, respectively. However, the pose estimation requires detecting these points in sub-pixel accuracy and their correspondences from the two views solved.

\begin{figure}[b]
    \centering
    \includegraphics[width=1\textwidth]{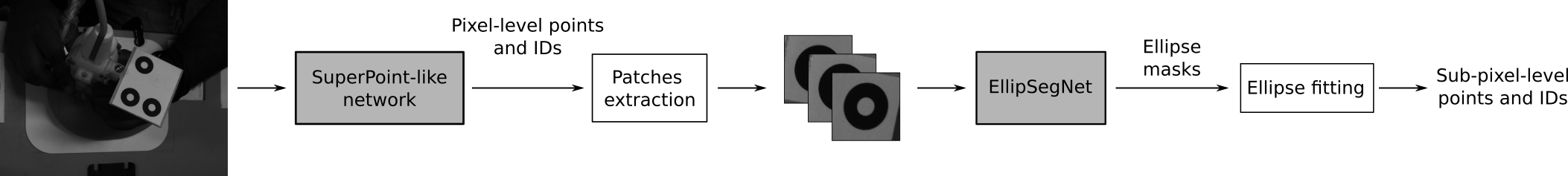}
    \caption{Sub-pixel circle centers detection and IDs classification pipeline. The gray shaded boxes correspond to convolutional neural networks.}
    \label{fig:detection_DL}
\end{figure}

\begin{figure}
    \centering
    \includegraphics[width=1\textwidth]{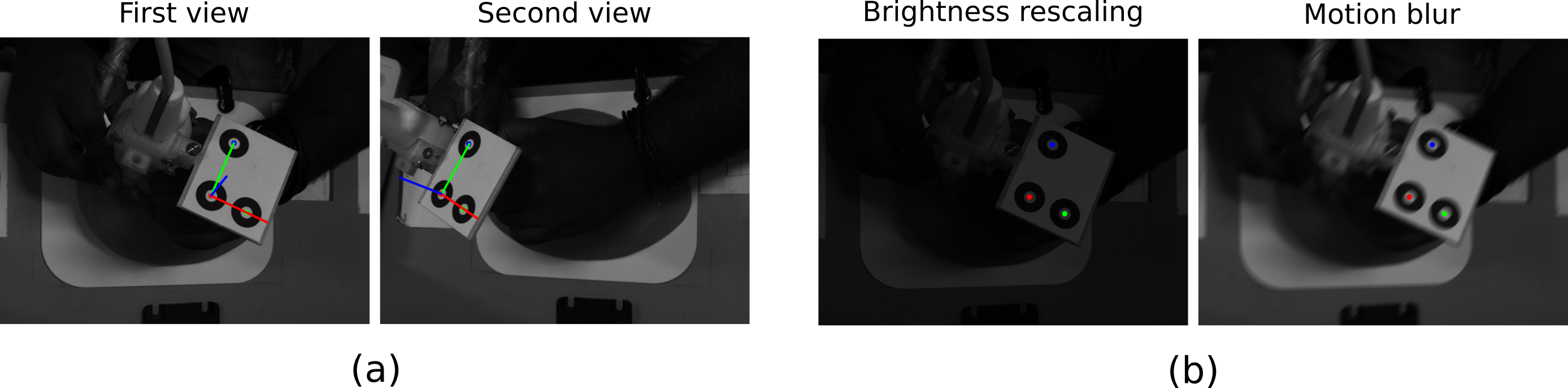}
    \caption{Example of the robustness of the target tracking method. (a) Target detection under synthetically degraded lighting and motion blur. (b) Final pose estimation with the stereo images.}
    \label{fig:roboust_detect}
\end{figure}

The pose estimation of the target is addressed in three stages using two CNNs. The first two stages consists of the sub-pixel centers detection of the target as shown Fig.~\ref{fig:detection_DL}. In the first stage, the centers are detected in a pixel-level accuracy, using a SuperPoint-like network proposed by Hu et al.~\cite{hu2019deep}. This network is a two-headed network based on the SuperPoint architecture~\cite{detone2018superpoint}, where an encoder is used to reduce the dimensionality of the input image. After the encoder, one of the heads estimates the probability that a circle center is present in each pixel (i.e., pixel-level detection). The other head classifies each point with a specific ID, solving the correspondence between the two views. In the second stage, with these rough detections, the algorithm extracts three patches centered on each circle center, such that the black circle is within the patch. As the circles project as ellipses, the detected contours are segmented with EllipSegNet, an encoder-decoder network. With the contour, the final sub-pixel center is estimated through ellipse fitting. Although it is known that the center of a projected circle and an ellipse deviate, for a small circle, such deviation is negligible~\cite{sun2019analysis}. In the last stage, the 3D coordinates of the three points are estimated through triangulation. With these points, the position and orientation of the target are calculated with respect to the $\{W\}$ frame. The translation vector $\mathbf{t}$ corresponds to the 3D coordinates of $c_0$ center. The rotation matrix can be established with the unit vectors of the target frame: $\hat{\mathbf{x}}$ calculated with $c_0$ and $c_1$ 3D centers, and $\hat{\mathbf{y}}$ calculated with $c_0$ and $c_2$ points. In this way the rotation matrix is $\mathbf{R} = [\hat{\mathbf{x}}, \;\: \hat{\mathbf{y}}, \;\: \hat{\mathbf{x}} \times \hat{\mathbf{y}}]$. Fig.~\ref{fig:roboust_detect}(a) shows an example of the acquired stereo images of the target and the final estimated target pose.

Both center detection networks were trained with real images. For the SuperPoint variant, a total of 5683 grayscale images were used. For the EllipSegNet, a total of 11010 patches were used for training. The center detection of the first two stages is robust to low lighting conditions, which is important for using the DLP projector in the SL system. We have to set a trade-off between the camera aperture and the exposure to avoid overexposed images for the SL system. This trade-off often leads to sub-optimal images for freehand US probe tracking, and the aperture cannot be modified to ensure the calibration remains the same. The tracking procedure is also robust to motion blur, crucial for real-time tracking of the US probe. Fig.~\ref{fig:roboust_detect}(b) shows an example of the detection under synthetic low lighting and synthetic motion blur applied. It is worth noting that it was trained and validated with more severe low-lighting and motion blur~\cite{meza2021markerpose}.

\subsection{3D Freehand Ultrasound Calibration}
\begin{figure}[b]
    \centering
    \includegraphics[width=0.52\textwidth]{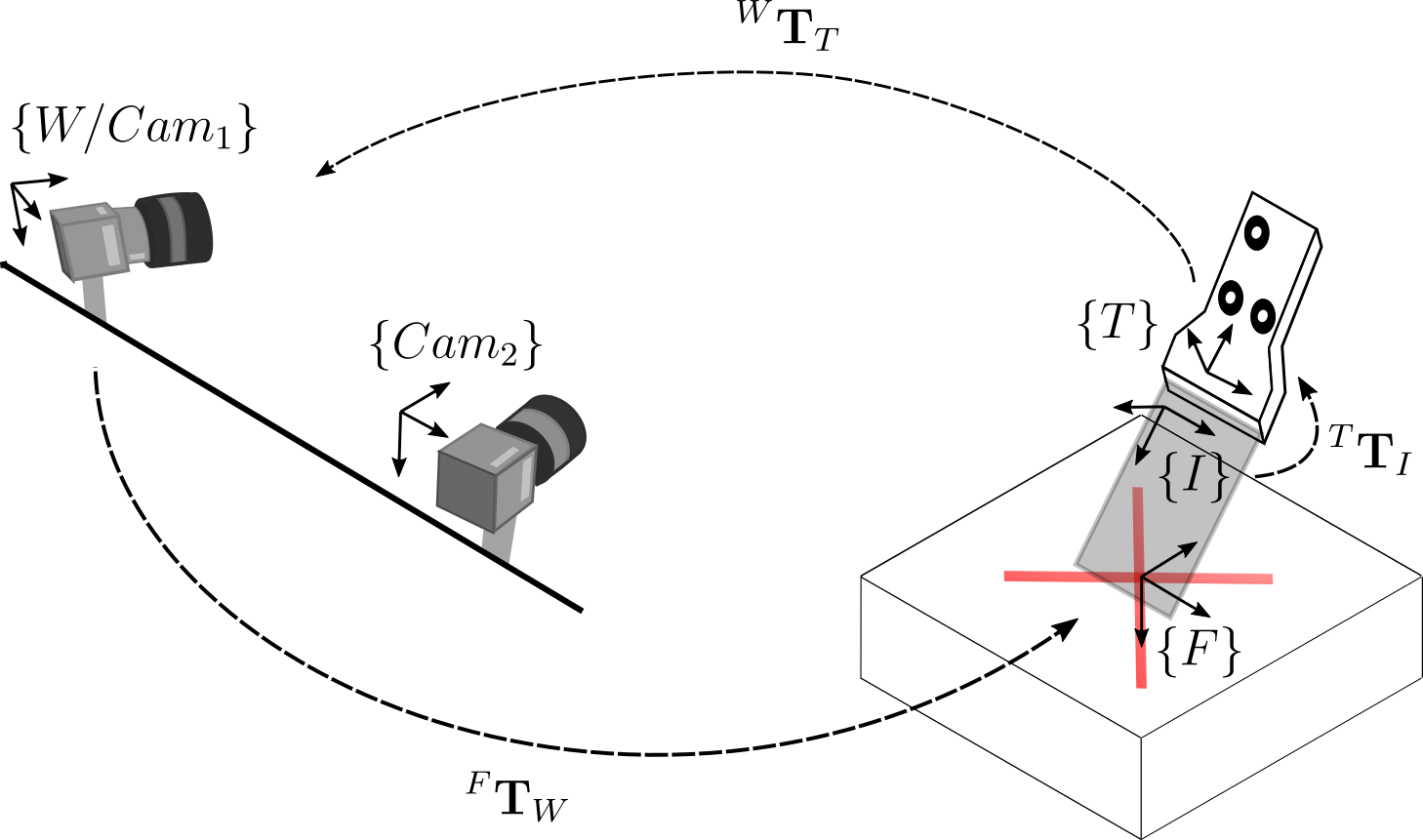}
    \caption{Transducer calibration: geometric relations and reference frames.}
    \label{fig:UScalib}
\end{figure}

Calibration of a 3D freehand US system requires tracking the probe's pose and consists of estimating the geometric relationship between the US image plane and the probe. For the calibration procedure, we used a point phantom of two cross-wires. In Fig.~\ref{fig:UScalib}, we illustrate the spatial relationships of the five coordinate systems involved in the probe calibration with our stereo vision system.

Using the transformation matrix $^W\mathbf{T}_T$ from the transducer's coordinate system $\{T\}$ to the world coordinate frame $\{W\}$ given by the stereo vision system, we need to estimate the transformation $^T\mathbf{T}_I$ from the image frame $\{I\}$ to the transducer's frame $\{T\}$. Furthermore, we also need to calculate the $x$ and $y$ scales of the US image $s_x$ and $s_y$ in millimeters per pixel to convert a pixel ($u$,~$v$) of the B-scan to metric units. The phantom coordinate system $\{F\}$ is placed so that its origin coincides with the cross-wires point. In this way, if we acquire a B-scan of the cross-wire phantom, we can map the segmented image coordinate of the point target in pixels to the physical phantom frame $\{F\}$ as
\begin{equation}
    [0\, 0\, 0\, 1]^T
    = \, ^F\mathbf{T}_W \, ^W\mathbf{T}_T \, ^T\mathbf{T}_I \,
    [s_x u\, s_y v\, 0\, 1]^T
    \enspace.
    \label{eq:calib_eqs}
\end{equation}
With this equation and a total of $N$ US images of the phantom, we can estimate the unknowns involved in the calibration.

To evaluate the transducer calibration quality, we measure the precision with the calibration reproducibility (CR) using the methods proposed by Prager et al.~\cite{prager1998rapid} and by Hsu et al.~\cite{hsu2009freehand}. Aspecific point ($u$,~$v$) in the B-scan is reconstructed from the image plane to the transducer coordinate system, and in this frame, we measure the variability of this point. Prager et al.~\cite{prager1998rapid}, used the bottom right pixel ($u_{\max}$,~$v_{\max}$) of the B-scan to measure CR under two different calibrations with the expression
\begin{equation}
    \mu_{CR1} = \left| ^T\mathbf{T}_{I1} \, \mathbf{X}^I_1 - \, ^T\mathbf{T}_{I2} \, \mathbf{X}^I_2 \right| \enspace,
    \label{eq:CR1}
\end{equation}
where $\mathbf{X}^I_i = [s_{xi}u_{max},\:  s_{yi}v_{max},\: 1,\: 0]^\mathsf{T}$.
We calculate CR with this expression using all possible pairs of calibration parameters, and finally, the mean of these values is reported as the final precision.

With the proposed strategy by Hsu et al.~\cite{hsu2009freehand}, calibration reproducibility is assessed as
\begin{equation}
    \mu_{CR2} = \frac{1}{N} \sum_{i=1}^N \left| ^T\mathbf{T}_{Ii} \, \mathbf{X}^I_i - \mathbf{\bar{X}}^T \right| \enspace,
    \label{eq:CR2}
\end{equation}
where $\mathbf{\bar{X}}^T$ is the mean in each dimension of all the reconstructed points $\mathbf{X}^I_i$ in the transducer $\{T\}$ coordinate system. As trial points, they used the center and the four corners of the image.

\subsection{Structured Light and Stereo Vision Calibration}
As our system is composed of two cameras and a projector, we need to calibrate the SL system for 3D reconstruction and the stereo vision system for tracking the US probe. As we can regard the projector as an inverse camera~\cite{zhang2006novel}, we calibrate both systems with the well-known stereo calibration procedure using a pinhole camera model~\cite{zhang2000flexible}. The $\{Cam_1\}$-$\{P\}$ and $\{Cam_1\}$-$\{Cam_2\}$ pairs are calibrated independently using the same pattern of asymmetric circles. For establishing the camera-projector correspondences, we used the phase-aided method proposed by Zhang and Huang~\cite{zhang2006novel} projecting fringe patterns onto the calibration board. To obtain a reliable phase-aided correspondence, we used binary defocused fringes of an 18-pixel pitch with the 18-step phase-shifting algorithm for phase recovery along with 7-bit gray coding patterns for absolute phase unwrapping~\cite{Zhang:2016tq}. To ensure high accuracy we acquired images from 34 poses of the calibration target.

After obtaining the $\{Cam_1\}$-$\{P\}$ and $\{Cam_1\}$-$\{Cam_2\}$ point correspondences, the calibration procedure is the same for both systems. We briefly describe the procedure for the SL system. The calibration consists of estimating the intrinsic and extrinsic parameters of the camera and the projector using the pinhole camera model for both devices, given as 
\begin{align}
s\mathbf{x} =  \mathbf{K} \mathbf{M} \mathbf{X} \enspace,
\label{eq:pinhole_model}
\end{align}
where $\mathbf{X}$ is a point in the 3D world coordinate system, $\mathbf{x}$ is its projection in the camera or projector sensor, $\mathbf{K}$ is the intrinsic parameter matrix, and $\mathbf{M}$ the extrinsic parameter matrix. In its expanded form, Eq.~\eqref{eq:pinhole_model} becomes
\begin{align} 
    \label{eq:projection}
    s \begin{bmatrix} u\\v\\1 \end{bmatrix}  = 
    \begin{bmatrix} 
    f_u & 0 & c_u\\
    0 & f_v & c_v\\
    0 & 0 & 1
    \end{bmatrix}
    \begin{bmatrix} 
    r_{11} & r_{12} & r_{13} & t_x\\
    r_{21} & r_{22} & r_{23} & t_y\\
    r_{31} & r_{32} & r_{33} & t_z
    \end{bmatrix} 
    \begin{bmatrix} 
    x^w\\
    y^w\\
    z^w\\
    1 
    \end{bmatrix} 
    \enspace,
\end{align}
where $f_u$ and $f_v$ are the focal length in the $u$ and $v$ direction respectively, ($c_u$,~$c_v$) is the principal point, $r_{ij}$ are the rotation matrix parameters, and $t_{i}$ the translation coefficients between the camera or projector and the world frame. As $\{Cam_1\}$ is also the world frame, the camera rotation matrix is the identity matrix, and the translation vector is the zero vector. Thus, for the camera and the projector, we have two sets of equations given by
\begin{align} 
    \label{eq:pinholeCam}
   s^c \mathbf{x}^c = \mathbf{K}^c \, \mathbf{M}^c \, \mathbf{X}^\text{w} \enspace,\\
    \label{eq:pinholeProj}
    s^p \mathbf{x}^p = \mathbf{K}^p \, \mathbf{M}^p \ \mathbf{X}^\text{w} \enspace.
\end{align}
With different views of the calibration pattern, we use Eqs.~\eqref{eq:pinholeCam}-\eqref{eq:pinholeProj} to solve for the camera's and projector's intrinsic parameters. Finally, we estimate the extrinsic parameters between the camera-projector pair. An analogous procedure is carried out for calibrating the stereo vision system formed by the two cameras. The obtained reprojection errors for camera-1, projector, and camera-2 are 0.1384, 0.1312, and 0.1508, respectively, which are quite small.

For 3D reconstruction with the SL system, we used an 8-step phase-shifting method with a center-line image for absolute phase-unwrapping to solve the phase correspondence. Eqs.~\eqref{eq:pinholeCam}-\eqref{eq:pinholeProj} are solved for obtaining the 3D surface.

\section{Experiments and Results}
We carried out different experiments to evaluate both the 3D freehand US and the SL systems separately and jointly as a multimodal technique.

\subsection{3D Freehand US Calibration Assessment}
With the $\{Cam_1\}$ - $\{Cam_2\}$ system calibrated, we carried out a total of five calibrations of the US probe with 30 images each one, using a linear transducer set to 5~cm depth. Table~\ref{tab:CR} shows the results of the calibration precision evaluated with $\mu_{CR1}$ and $\mu_{CR2}$, using Eqs.~\eqref{eq:CR1}- \eqref{eq:CR2}, respectively. We use five trial points for the CR evaluation for both metrics, the center, and the four image corners, where our US image is $321 \times 408$~px. Our results are similar to those previously reported. For example, Hsu et al.,~\cite{hsu2009freehand} report a CR of 0.27~mm with $\mu_{CR2}$, measured in the center of the image and using a point phantom and the probe at 3~cm depth. Furthermore, Lindseth et al.~\cite{lindseth2003probe} report a CR at the center of the B-scan of 0.62~mm with a point phantom and a linear probe at 8~cm depth using $\mu_{CR1}$. Finally, we report an RMS error of all the equations obtained through the five calibrations of 0.4231~mm.

\begin{table}[h]
\centering
\begin{tabular}{lcc}
\hline
\multicolumn{1}{c}{\textbf{Trial point}} & $\mu_{CR1}$ (mm) & $\mu_{CR2}$ (mm) \\ \hline
Center                                     & 0.4337         & 0.2736    \\ 
Bottom right ($u_{max}$, $v_{max}$)        & 0.5513         & 0.3453    \\ 
Mean (center and four corners)             & 0.5905         & 0.3681    \\
\hline
\end{tabular}

\caption{Precision assessment: calibration reproducibility results with a total of 5 calibrations.}
\label{tab:CR}
\end{table}

\subsection{Structured Light Assessment}
To evaluate the calibration of the SL system, we reconstructed a plane in five different poses. The volume covered with the planes is $252.97 \times 198.56 \times 107.41$~mm. The RMS errors between the reconstructed and the ideal plane estimated through least-squares is shown in Table~\ref{tab:rms}.

\begin{table}[h]
\centering
\resizebox{\textwidth}{!}{%
\begin{tabular}{lccccc}
\hline
Pose           & 1                     & 2                     & 3                     & 4                     & 5                     \\ \hline
RMS error (mm) & 0.1161                & 0.1244                & 0.1305                & 0.1120                & 0.1006                \\
$x$ range (mm) & {[}-116, 126.07{]}    & {[}-120.58, 129.91{]} & {[}-116.4, 132.39{]}  & {[}-119.09, 130.12{]} & {[}-116.26, 129.52{]} \\
$y$ range (mm) & {[} -96.75, 90.41{]}  & {[} -99.97, 93.57{]}  & {[}-101.23, 97.33{]}  & {[} -99.85, 88.38{]}  & {[} -99.39, 84.78{]}  \\
$z$ range (mm) & {[} 659.89, 680.09{]} & {[} 682.8, 703.48{]}  & {[} 675.58, 713.57{]} & {[} 642.23, 701.56{]} & {[} 606.16, 697.98{]} \\ \hline
\end{tabular}%
}
\caption{3D reconstruction assessment results under different poses of a flat board.}
\label{tab:rms}
\end{table}

We also measured a sphere with a 19.8~mm radius to evaluate the system calibration with a different geometry. The estimated radius with the 3D reconstruction is 19.755~mm, which gives an absolute error of 0.045~mm. Finally, the RMS error between the reconstructed and the ideal sphere is 0.0399~mm.

\subsection{Multimodal System Assessment}
\begin{figure}
    \centering
    \includegraphics[width=1\textwidth]{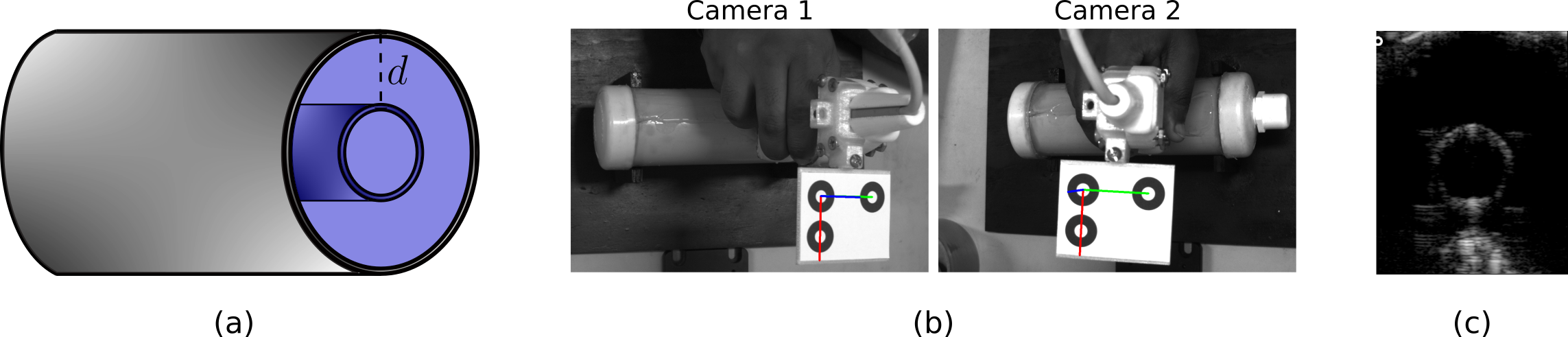}
    \caption{Validation experiment of the multimodal system with two concentric cylinders. (a) Diagram of the arrangement of cylinders and the measured distance $d$ between the internal and external reconstruction. (b) Captured stereo images during ultrasound acquisition of the internal cylinder. (c) B-scan example of the internal cylinder.}
    \label{fig:cylin_exp}
\end{figure}

We evaluated the proposed system as a multimodal technique through two experiments. For a quantitative evaluation, we designed an experiment where we measured two concentric cylinders. Fig.~\ref{fig:cylin_exp}(a) shows a schematic of the experiment, where the inner cylinder is submerged in water to be measured with 3D freehand US, while the outer cylinder is reconstructed with SL. This experiment aims to estimate the distance $d$ between the inner and outer cylinder, i.e., between the 3D freehand US reconstruction and the SL reconstruction. Fig.~\ref{fig:cylin_exp}(b) shows the stereo images of the measured object during US acquisition, and Fig.~\ref{fig:cylin_exp}(b) the corresponding US image. The outer diameter of the internal cylinder is 15.82~mm, of the external cylinder is 60~mm, and the distance $d$ between the cylinders is 22.38~mm.

\begin{figure}
    \centering
    \includegraphics[width=1\textwidth]{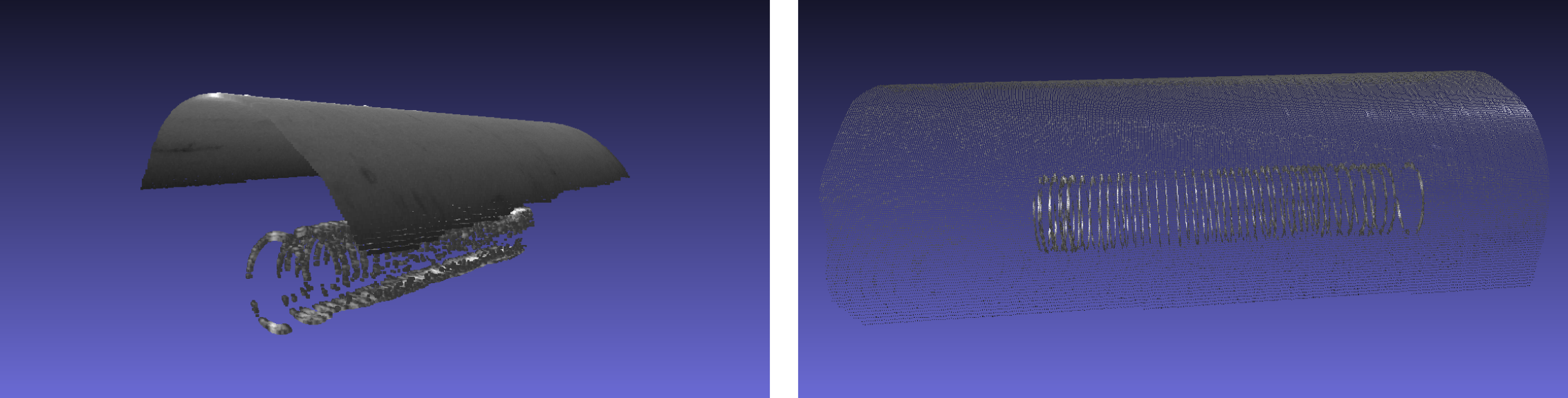}
    \caption{Internal and external reconstruction results of the concentric cylinders in the same coordinate system used for the validation experiment.}
    \label{fig:cylin_recons}
\end{figure}

For this experiment, a total of 94 US slices were captured along the cylinder. The multimodal reconstruction results are shown in Fig.~\ref{fig:cylin_recons}, where we have the segmented rings from the US images mapped to the 3D space and the external cylindrical cap reconstructed with the SL system. With the point cloud obtained with each technique, we estimated the least-squares ideal cylinder. The internal cylinder's estimated diameter with the freehand US reconstruction is 15.1~mm which gives an absolute error of 0.72~mm. The external diameter of the outer cylinder is 60.10~mm with 0.1~mm of absolute error. Finally, the estimated $d$ distance between both reconstructions is 22.50~mm with an absolute error of 0.12~mm. These results show an adequate performance of the proposed multimodal 3D imaging technique.

\begin{figure}
    \centering
    \includegraphics[width=1\textwidth]{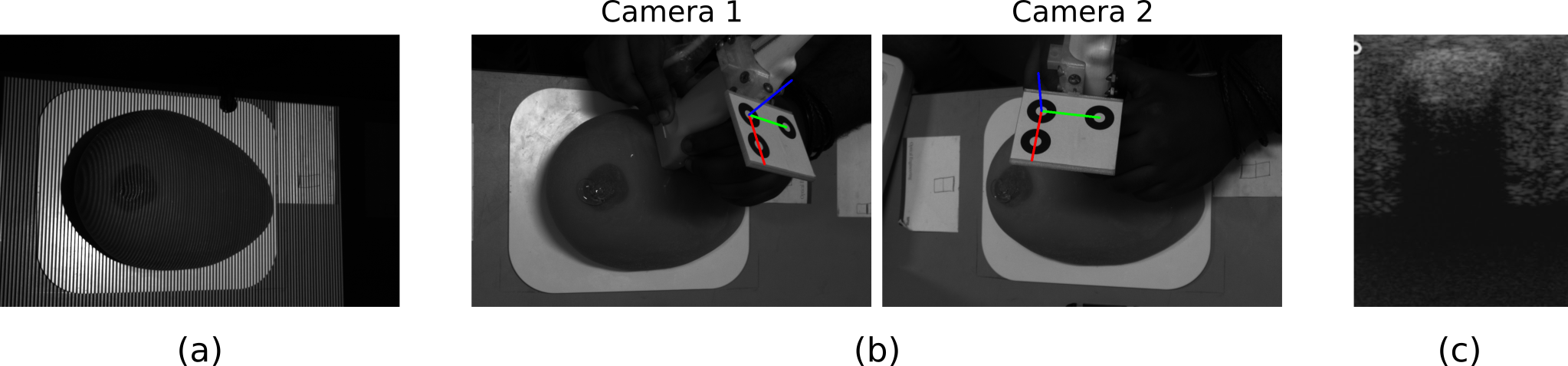}
    \caption{Structured light and freehand US acquisition of the phantom for the qualitative experiment. (a) Projected pattern acquisition. (b) Stereo images with estimated target pose for freehand US. (c) US image}
    \label{fig:SL_US_acq_phatom}
\end{figure}

\begin{figure}
    \centering
    \includegraphics[width=1\textwidth]{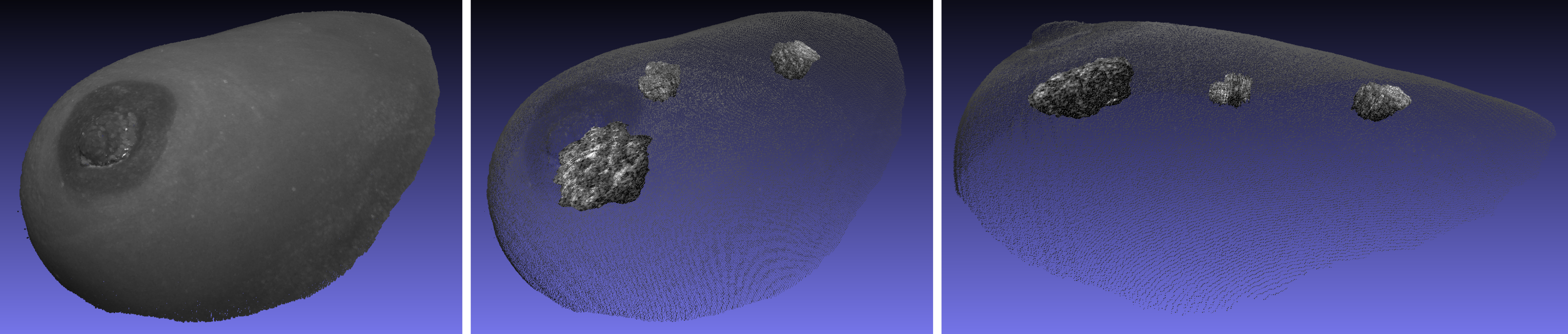}
    \caption{Qualitative results of the multimodal reconstruction. The 3D surface of the breast phantom from the SL technique, and the three tumors of the phantom reconstructed with 3D freehand ultrasound.}
    \label{fig:US_FPP}
\end{figure}

The second experiment consisted of evaluating the proposed multimodal system using a breast phantom 3B SONOtrain P125 made from US material with three tumors. $170 \times 130 \times 55$ mm are the phantom's dimensions, and the tumors have three different sizes: 27~mm, 14~mm, and 12~mm diameters. We reconstructed the external breast surface (external features) with the SL technique and the three tumors (internal features) with the 3D freehand US method. Thus, we projected the SL patterns for surface 3D reconstruction as shown Fig.~\ref{fig:SL_US_acq_phatom}(a). Then, we acquired US slices of the three tumors for the 3D mapping. An example of the pose estimation with the stereo vision system is shown in Fig.~\ref{fig:SL_US_acq_phatom}(b), and the B-scan image simultaneously acquired with the stereo images is shown in Fig.~\ref{fig:SL_US_acq_phatom}(c). Fig.~\ref{fig:US_FPP} shows the result of both reconstructions relative to the $\{W\}$. Both reconstructions are in the same coordinate system, and the reconstructed tumors are totally within the phantom surface. The result is in agreement with the expected position of the tumors.

\section{Potential Applications of the Multimodal System}
\begin{figure}[b]
    \centering
    \includegraphics[width=0.8\textwidth]{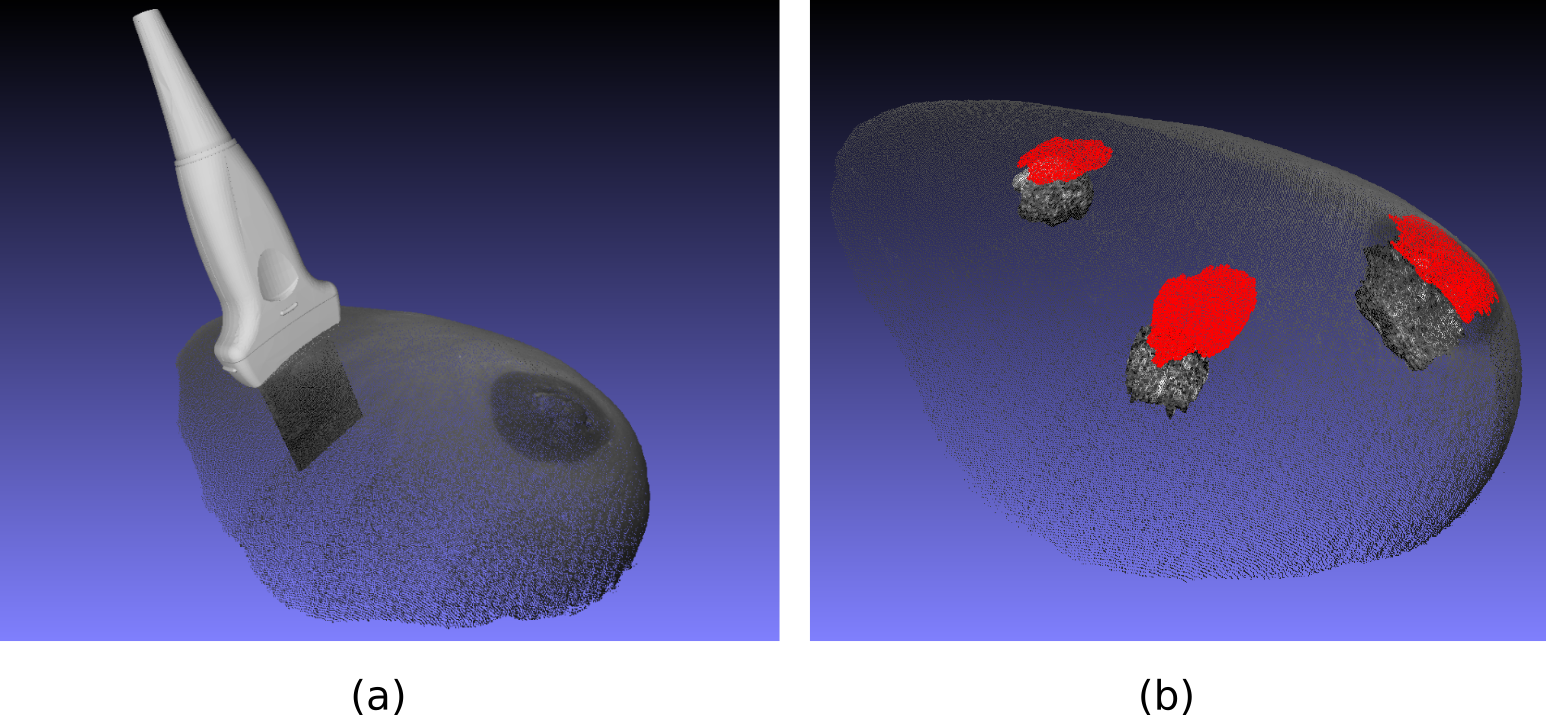}
    \caption{Potential applications of the proposed 3D multimodal imaging system. (a) The proposed multimodal technique as a navigation system. The tracked ultrasound probe mapped in 3D with the ultrasound scan and the 3D surface. (b) The projector as an active device to project the internal structures in the external surface.}
    \label{fig:navig}
\end{figure}

As we track the US probe's pose in the same coordinate system of the 3D reconstructions, we can use the proposed multimodal system as a navigation system where the probe can be displayed in real-time with the reconstructions as shown in Fig.~\ref{fig:navig}(a). In this way, we have the US slice mapped in 3D and the probe with the 3D surface reconstruction. This visualization can be useful for image interpretation and spatial understanding during surgery. Additionally, we can track any other instrument by attaching a marker to display it in 3D with the other 3D reconstructions, e.g., a needle for biopsy procedures.

Furthermore, the DLP projector can be used as an active device to project the internal reconstructed structures onto the subject's skin. In a similar fashion to fluorescence-guided surgery, the internal structures are highlighted by projecting light onto the tissue. For example, in breast cancer biopsy~\cite{schaafsma2013clinical}. We can also display with the projector the location of the internal 3D structures. Fig.~\ref{fig:navig}(b) shows a simulated example, where the reconstructed tumors are highlighted on the breast's surface.

\section{Summary}

We proposed a low-cost 3D multimodal medical imaging technique by combining 3D freehand ultrasound and structured light. With this multimodal technique, it is possible to complement the internal information acquired with 3D freehand ultrasound with the external surface obtained from structured light. Both reconstructions are in the same global coordinate system avoiding complicated data registration procedures. Furthermore, using a pose estimation system based on Convolutional Neural Networks and stereo vision avoids using costly commercial tracking systems. The experimental results show the proposed system's high accuracy and its suitability for preoperative or intraoperative procedures such as surgical planning or guidance.

\subsection* {Acknowledgments}
This work has been partly funded by Universidad Tecnológica de Bolívar project C2018P005. J.~Meza thanks Universidad Tecnológica de Bolívar for a post-graduate scholarship and MinCiencias, and MinSalud for a ``Joven Talento" scholarship. Parts of this work were presented at the 15th International Symposium on Medical Information Processing and Analysis~\cite{Meza:2020ge}.

\subsection* {Code, Data, and Materials Availability}
The software implementation is available at \url{https://github.com/jhacsonmeza/StructuredLight_3DfreehandUS}. The image acquisition software was developed in C++ and the proposed method for optical tracking and the SL 3D reconstruction was implemented in Python using OpenCV 4.4.0. The deep learning model was implemented with PyTorch 1.6.0. In the current implementation the data is acquired and processed off-line for visualization. A real-time implementation is currently being developed.

\bibliography{report}   
\bibliographystyle{spiejour}   

\end{spacing}
\end{document}